\documentclass[10pt,twocolumn,letterpaper]{article}

\usepackage[pagenumbers]{iccv} 

\definecolor{iccvblue}{rgb}{0.21,0.49,0.74}
\usepackage[pagebackref,breaklinks,colorlinks,allcolors=iccvblue]{hyperref}
\usepackage{url}
\usepackage{amsmath} 
\usepackage{graphicx}
\usepackage{booktabs}
\usepackage{algorithm}
\usepackage{algpseudocode}

\usepackage{algorithm}
\usepackage{algpseudocode}
\usepackage{multirow}
\usepackage{multicol}
\usepackage{colortbl}

\newcommand{\methodname}{WoTE}
\usepackage[utf8]{inputenc} 
\usepackage[T1]{fontenc}    
\usepackage{hyperref}       
\usepackage{url}            
\usepackage{booktabs}       
\usepackage{amsfonts}       
\usepackage{nicefrac}       
\usepackage{microtype}      
\usepackage{xcolor}         
\usepackage{wrapfig} 
\usepackage{makecell}
\usepackage{xcolor}
\usepackage{placeins}
\usepackage{afterpage}
\usepackage{marvosym}

\def\paperID{4321} 
\def\confName{ICCV}
\def\confYear{2025}

\title{End-to-End Driving with Online Trajectory Evaluation via BEV World Model}

\author{
  Yingyan~Li$^{*}$ \quad Yuqi~Wang$^{*}$ \quad Yang~Liu \quad Jiawei~He  \quad  Lue~Fan\textsuperscript{\Letter}  \quad
  Zhaoxiang~Zhang\textsuperscript{\Letter}\\
  NLPR, Institute of Automation, Chinese Academy of Sciences (CASIA) \\
  University of Chinese Academy of Sciences (UCAS) \\ 
}

\begin{document}
\maketitle
\footnotetext[1]{*: Co-first author, \textsuperscript{\Letter}: Co-advisor. Email: liyingyan2021@ia.ac.cn}
\begin{abstract}
End-to-end autonomous driving has achieved remarkable progress by integrating perception, prediction, and planning into a fully differentiable framework. 
Yet, to fully realize its potential, an effective online trajectory evaluation is indispensable to ensure safety.
By forecasting the future outcomes of a given trajectory, trajectory evaluation becomes much more effective.
This goal can be achieved by employing a world model to capture environmental dynamics and predict future states.
Therefore, we propose an end-to-end driving framework \textbf{\methodname{}}, which leverages a BEV \textbf{Wo}rld model to predict future BEV states for \textbf{T}rajectory \textbf{E}valuation.
The proposed BEV world model is latency-efficient compared to image-level world models and can be seamlessly supervised using off-the-shelf BEV-space traffic simulators.
We validate our framework on both the NAVSIM benchmark and the closed-loop Bench2Drive benchmark based on the CARLA simulator, achieving state-of-the-art performance.
Code is released at \url{https://github.com/liyingyanUCAS/WoTE}.
\end{abstract}    
\section{Introduction}
\label{sec:intro}

\begin{figure}[!]
    \centering
    \includegraphics[width=\linewidth]{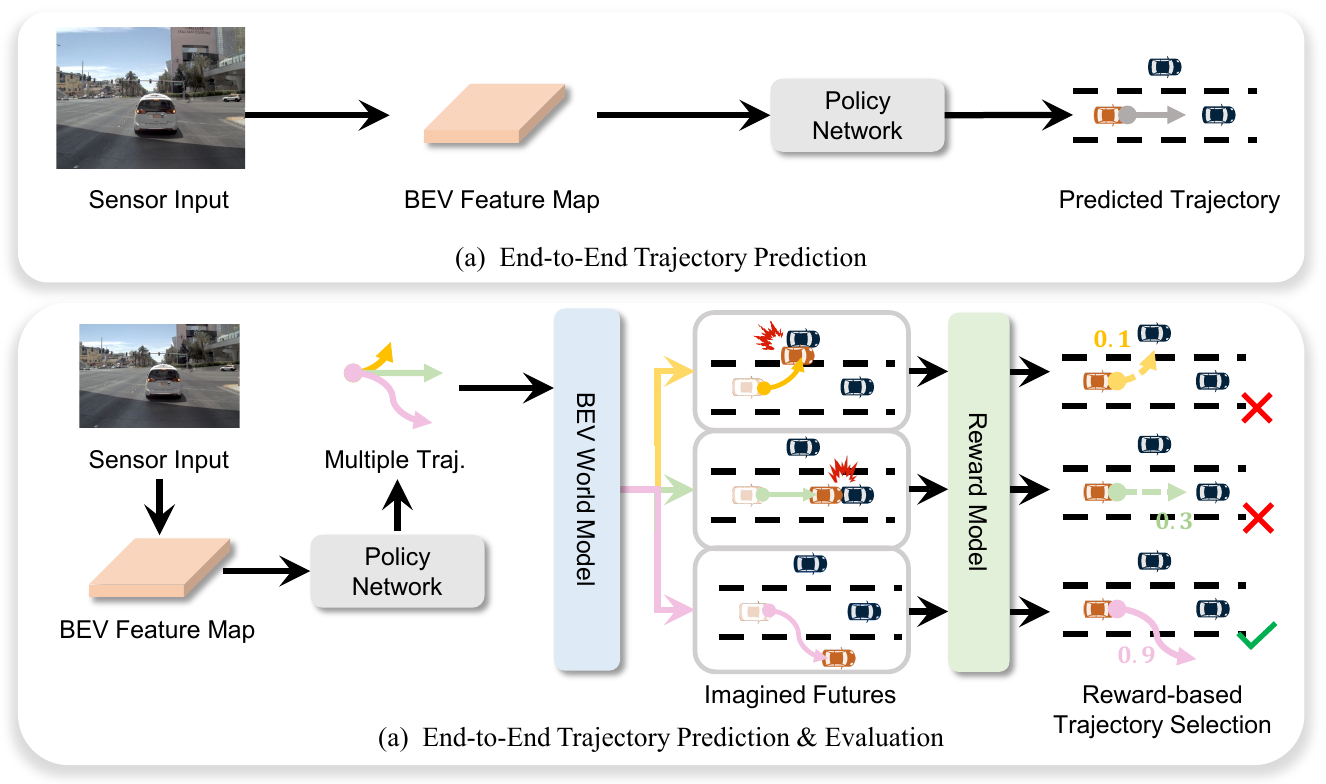}
    \caption{\textbf{The illustration of end-to-end trajectory evaluation with BEV world model.} (a) Previous end-to-end driving methods primarily focus on learning a high-quality trajectory. (b) Our \methodname{} consists of end-to-end trajectory prediction and evaluation. First, we predict multi-modal trajectories. Next, given these trajectories, we leverage the BEV world model to predict the corresponding future BEV states. The reward model evaluates these trajectories based on the predicted states and assigns rewards, selecting the highest-reward trajectory.}
\label{fig:fig1}
\end{figure}

End-to-end autonomous driving~\cite{law, tampuu2020survey, prakash2021multi, chen2024end} has achieved remarkable success in recent years and garnered widespread attention.
While most approaches~\cite{hu2023_uniad, jiang2023vad, hu2022stp3} primarily focus on predicting a high-quality trajectory, 
recent works~\cite{vadv2, li2024hydra} suggest that predicting multiple trajectories simultaneously better reflects the multi-modal nature of driving and leads to improved performance. 
Given these multiple possible trajectories, it becomes crucial to effectively \emph{evaluate} them in order to ensure the safety and reliability of end-to-end autonomous driving systems.

Traditional trajectory evaluation methods are rule-based~\cite{treiber2000congested, fan2018baidu, hu2022st, dauner2023parting} and rely on perception results, such as bounding boxes and maps, to assess trajectories. 
These methods are sensitive to perception inaccuracies and may not be directly optimized in an end-to-end manner. 
Recently, some end-to-end driving approaches~\cite{li2024hydra, vadv2} have incorporated trajectory evaluation.
Their evaluations rely on the current state. 
However, evaluating a trajectory candidate is particularly challenging without knowledge of the \emph{future} led by this trajectory. 
Similar to human drivers, who anticipate futures before making decisions, a trajectory evaluation network would benefit from predicting the corresponding future states of trajectory candidates.

To effectively predict future states, it is intuitive to adopt methodologies from reinforcement learning, where world models have successfully facilitated future state predictions.
In reinforcement learning, model-based methods~\cite{henaff2018modelpredictive, chen2021learning, qiao2025modelling} have consistently shown superior performance compared to model-free approaches~\cite{mnih2015human, haarnoja2018soft}, largely due to their ability of modeling the environmental dynamics and reasoning about future scenarios through world models. 
Inspired by this success, we propose adopting a world model~\cite{hu2023gaia,gao2024vista,wang2023drivedreamer,wang2024driving, wang2024drivingdojo} within end-to-end autonomous driving frameworks to facilitate accurate prediction of future states, thus significantly improving trajectory evaluation.

However, leveraging the world model for trajectory evaluation presents some technical issues. 
First, a good representation of future scenarios is needed. 
Many existing world models for driving~\cite{hu2023gaia,wang2023drivedreamer,wang2024driving,wang2024freevs,yan2024streetcrafter} predict the images of future scenarios with diffusion models~\cite{rombach2022diffusion}, which can be time-consuming and not suitable for real-time driving. 
Although another line of work~\cite{fan2024freesim, zhou2025flexdrive, zhao2024drivedreamer4d,zhao2025recondreamer++} for driving simulation can render free-viewpoint future videos in real-time, they need off-board reconstruction and cannot be adopted in our onboard driving setting.
Second, we lack supervision for future states.
Supervising the predicted future states of the world model is difficult because world models need to imagine multiple future states based on the multiple trajectory candidates, while only one future state is available in the real-world datasets.

To address these technical issues, we propose \methodname{} as shown in Fig.\ref{fig:fig1}. 
We propose to predict future states in the BEV space.
The information in BEV space is much more compact than in raw images, which enables a more efficient single-step feed-forward future prediction than the time-consuming multi-step denoising in previous driving world models~\cite{hu2023gaia, wang2023drivedreamer, wang2024driving, zhang2024bevworld}.
This characteristic of BEV space makes \methodname{}  highly suitable for real-time driving applications. 
Additionally, the predicted future BEV states can be easily supervised using off-the-shelf BEV-space traffic simulators such as nuPlan~\cite{nuplan, gulino2024waymax}. 
These simulators model the scenario in BEV space by representing the map and objects using BEV semantic maps, creating a dynamic simulation of the driving environment. 
Consequently, we can use the BEV semantic maps from these simulated scenarios as supervision for the prediction of the BEV world model. 
Moreover, these simulators are equipped with well-established evaluation rules to assess the simulated future BEV scenarios, which provide target rewards.

We validate our framework on both NAVSIM~\cite{navsim} benchmark and closed-loop Bench2Drive~\cite{jia2024bench2drive} benchmark based on the CARLA~\cite{dosovitskiy2017carla} simulator and achieve state-of-the-art performance.
Our experiments show that incorporating future states into trajectory evaluation is essential for improved performance.
Additionally, the trajectory evaluation module, built on the world model shows great generalization ability across diverse trajectories.
Our contributions are as follows: 
\begin{itemize} 
\item We highlight the importance of utilizing imagined future states for trajectory evaluation in end-to-end autonomous driving. 
\item We introduce \methodname{}, which consists of an end-to-end trajectory evaluation module based on a BEV world model. The world model in BEV space enables efficient real-time future prediction and overcomes the scarcity of multi-future supervisions.
\item \methodname{} is validated on both the NAVSIM~\cite{navsim} benchmark and the closed-loop CARLA-based Bench2Drive~\cite{jia2024bench2drive} benchmark, achieving state-of-the-art performance.
\end{itemize}

\section{Related Works}
\subsection{End-to-end Autonomous Driving}
End-to-end autonomous driving~\cite{hu2022uniad, jiang2023vad, vadv2, yuan2024drama, jiang2025alphadrive, sima2025centaur, liao2024diffusiondrive, gao2025rad, hydra-next, law} is to train a model that directly maps sensor inputs to trajectories or control signals instead of decomposing the driving task into traditional sub-tasks.
These works can be divided into two categories according to the training paradigm: imitation-learning-based and reinforcement-learning-based. 
Imitation learning is the most common practice in end-to-end autonomous driving. 
Their predicted trajectories are supervised by the expert trajectories.
For instance, P3~\cite{sadat2020p3}, takes a differentiable semantic occupancy as the intermediate representation and learns actions from human trajectories.
Following this, ST-P3~\cite{hu2022stp3} learns spatial-temporal features for perception, prediction and planning tasks simultaneously.
UniAD~\cite{hu2022uniad} proposed goal-oriented planning, combining and leveraging advantages of perception and motion prediction modules. 
VAD~\cite{jiang2023vad} introduces vectorized representations for end-to-end planning.
There are also some reinforcement-learning-based methods.
For example, MaRLn~\cite{marln} utilizes implicit affordances to develop a reinforcement learning algorithm. 
LBC~\cite{chen2020lbc} trained a vision-based sensorimotor agent from the privileged agent.
Combining imitation learning and reinforcement learning, Hydra-MDP~\cite{li2024hydra} utilizes knowledge distillation to obtain supervisions from both rule-based planners and human drivers. 
It is a promising research direction as more supervisions are utilized.

\subsection{Trajectory Evaluation}
Trajectory evaluation is essential for ensuring the reliability and safety of autonomous driving systems. 
We divide the main trajectory evaluation methods into two categories: model-free and model-based. 
Model-free methods directly assess the learned policy using observed data without explicitly modeling environmental dynamics. 
For instance, Monte Carlo policy evaluation~\cite{arnold2022policy} uses the empirical mean return as the metric. 
The value function approximates the expectation based on sampled returns. 
Temporal-difference learning~\cite{dann2014policy} uses bootstrapping, where the current estimate of the value function is used to generate target values for updating the value function.
Subsequently, model-based trajectory evaluation methods have gained increasing popularity in recent years. 
By leveraging learned world models that capture environmental dynamics, these methods enable more accurate and predictive policy evaluation.
For example, PPUU~\cite{henaff2018modelpredictive} measures policy performance by quantifying uncertainty in predicted future states. Rails~\cite{chen2021learning} evaluates trajectories through a tabular dynamic programming mechanism, and MARL-CCE~\cite{qiao2025modelling} employs learned environmental dynamics to assess multi-agent policies in driving scenarios.
However, these methods often rely on explicit trajectory representations and non-differentiable metrics, which limits their end-to-end optimization capabilities. 
In contrast, our approach evaluates trajectories using BEV features and encoded trajectory embeddings, enabling fully differentiable and thus end-to-end trajectory evaluation.

\begin{figure*}[!t]
\centering
\includegraphics[width=\textwidth]{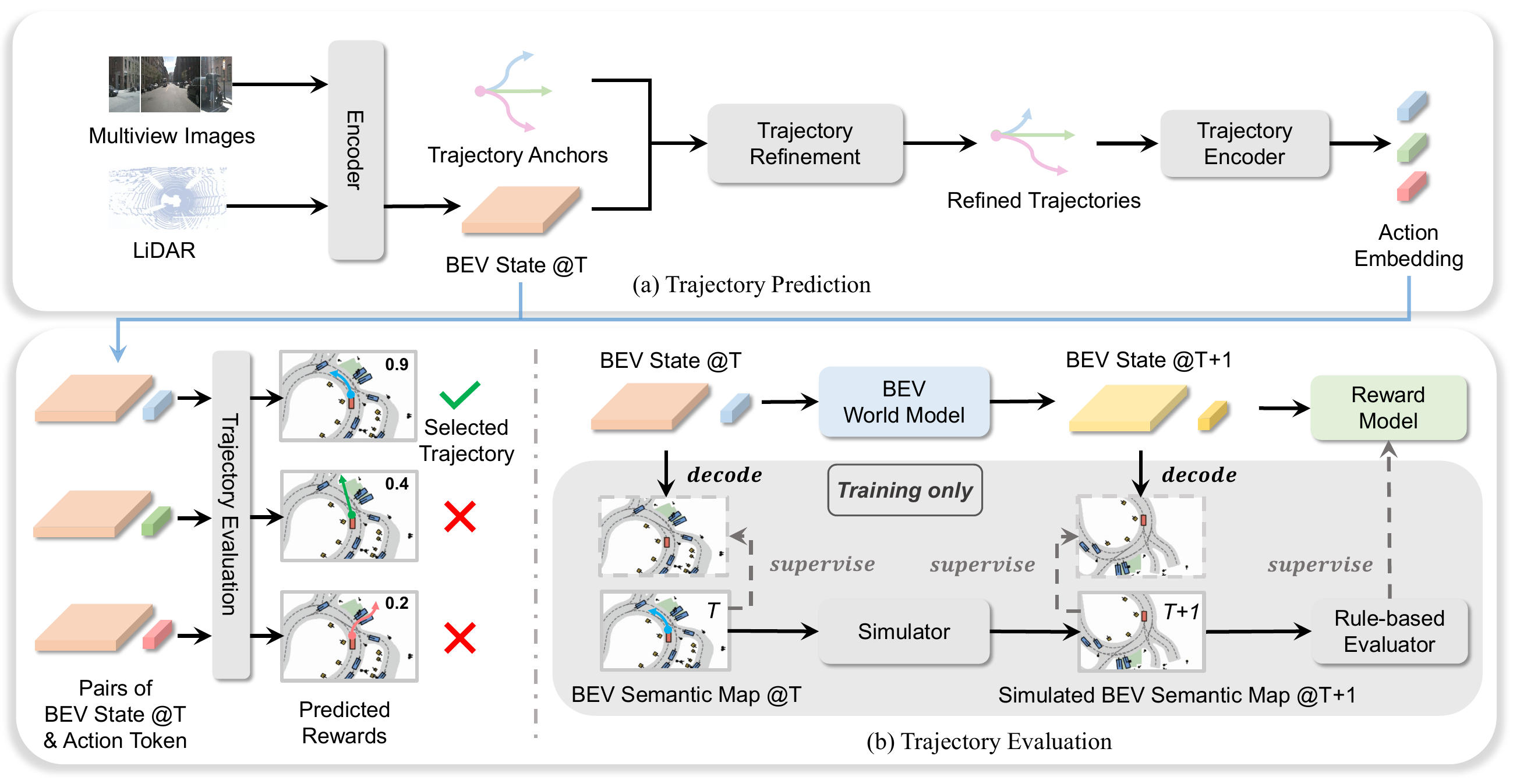}
\caption{\textbf{Overall structure of \methodname{}}. Our model predicts and evaluates trajectory in an end-to-end manner. 
It is mainly divided into two parts.
(a) Trajectory Prediction: the BEV encoder encodes the multi-modal observations into a BEV feature map (state) and proposes multiple trajectory candidates. (b) Trajectory Evaluation: 
Given the current BEV state and trajectory candidates, we leverage the BEV World Model to predict the corresponding future BEV states of these trajectories.
The Reward Model then predicts rewards with the help of future BEV states.
The trajectory with the highest reward is selected as the final trajectory.
}
\label{fig:pipeline}
\end{figure*}

\section{Method}
In this section, we introduce four key components of \methodname{}: Trajecoty Predictor (\cref{sec:policy}), BEV World Model (\cref{sec:bevwm}), Reward Model (\cref{sec:reward_predictor}) and BEV Space Supervision~(\cref{sec:bev_space_supervision}). 
As illustrated in Figure~\ref{fig:pipeline}, our method integrates future state prediction via the BEV World Model and reward prediction via the Reward Model within a unified framework.
Specifically, the Trajecoty Predictor encodes the multi-modal observations into BEV state, represented as a BEV feature map, and provides multiple trajectory proposals. The BEV World Model predicts corresponding future BEV states based on these trajectory proposals. With the help of future states, the Reward Model predicts the reward of each trajectory. Finally, we choose the trajectory with the highest reward as the final trajectory. 

\subsection{Trajectory Predictor}
\label{sec:policy}
As shown in Figure~\ref{fig:pipeline} (a), the Trajectory Predictor aims to predict multiple trajectory candidates. 
First, the BEV encoder encodes the multi-modal sensor inputs into a unified BEV feature map. 
Then, given trajectory anchors, the trajectory refinement module refines trajectories based on the BEV feature map. 
We will now describe each part in detail.

\paragraph{Multi-modal BEV Encoder.}
Following TransFuser~\cite{transfuser}, our method leverages multi-modal sensor inputs, including LiDAR and multi-view RGB images~\cite{fsf, dcd}.  
We use a BEV encoder~\cite{transfuser} to transform the current multi-modal inputs into a unified BEV feature map $\mathbf{B} \in \mathbb{R}^{h \times w \times c}$. 
For convenience in the following text, we refer to the BEV feature map as the BEV state.
$h$ and $w$ represent the height and width of the BEV state, while $c$ denotes the number of channels.

\paragraph{Trajectory Anchors.} 
Following methods in~\cite{li2024hydra, vadv2}, 
we generate trajectory anchors $\mathbf{\tau}$ by K-Means clustering. 
This clustering is performed on all the expert trajectories collected from the training dataset.
The number of trajectory anchors is denoted as $N$.

\paragraph{Trajectory Refinement.}
As illustrated in Figure~\ref{fig:pipeline} (a), the trajectory refinement module takes the trajectory anchors and the current BEV states as inputs to predict the refined trajectories. Specifically, we first use a trajectory encoder $\mathbf{TE}$, which is implemented by an MLP, to encode the trajectory anchors \(\mathbf{\tau}\) into feature vectors. Then, these feature vectors are used as queries to perform cross attention with the current BEV state \(\mathbf{B}\), which serves as both the key and value. The output of the cross-attention is then fed into an MLP head to predict the offsets. The offset is added to the trajectory anchors to produce the refined trajectories. 
The entire process can be expressed as
\begin{equation}\label{eq:traj_refine}
   \mathbf{\hat{\tau}} = \mathbf{\tau} + \text{MLP}(\text{CrossAttention}(\mathbf{TE}(\mathbf{\tau}), \mathbf{B}, \mathbf{B})),
\end{equation}
where $\mathbf{\hat{\tau}}$ is the refined trajectories.

\subsection{BEV World Model} 
\label{sec:bevwm}
In this section, we use the BEV World Model to predict the future BEV states, as illustrated in Figure~\ref{fig:pipeline} (b). 
The process is divided into two steps.
First, the current BEV state and refined trajectories are combined to form the inputs for the world model. 
Next, the world model recurrently predicts future states over multiple time steps.

\paragraph{Input of World Model.}  
Given \( N \) refined trajectories \(\mathbf{\hat{\tau}}\) and the current BEV state \(\mathbf{B}_t\) at time $t$, we construct \( N \) state-action pairs as the inputs of world model.  
We use the trajectory encoder $\mathbf{TE}$ in Eq.~\ref{eq:traj_refine}, with shared parameters, to encode \(\mathbf{\hat{\tau}}\) into action embeddings \(\mathbf{A}_t\).  
Given \(\mathbf{A}_t = \{\mathbf{a}^1_t, \mathbf{a}^2_t, \dots, \mathbf{a}^N_t\}\), each state-action pair is represented as \((\mathbf{B}_t, \mathbf{a}^i_t)\), where $i \in \{1, \dots, N\}$.

\paragraph{Recurrent Future Prediction.}  
Given the state-action pairs, we use a world model to predict their corresponding future states.
In contrast to prior work~\cite{wang2024driving}, which predict the future in image space, we build a world model in BEV space as it is much more efficient. 
In detail, given a state-action pair \((\mathbf{B}_t, \mathbf{a}^i_t)\), we flatten $\mathbf{B}_t$ into \( hw \) feature vectors. The dimension of each vector is \( c \). Then we concatenate these vectors with the action embedding $\mathbf{a}^i_t$, resulting in a total of \( h w + 1 \) feature vectors, namely $\mathbf{F}_i\in \mathbb{R}^{(hw+1)\times c}$. 
$\mathbf{F}_i$ are then fed into the world model,  whose network architecture is a transformer encoder,  as $\mathbf{F}_{i+1} = \text{TransformerEncoder}(\mathbf{F}_i).$
The output of the world model consists of \( h \times w + 1 \) feature vectors, corresponding to the predicted future states \(\mathbf{B}_{t+1}^i\) and future actions \(\mathbf{a}^i_{t+1}\).
In summary, we can formulate this process in the following equation
\begin{equation}
  (\mathbf{B}_{t+1}^i, \mathbf{a}^i_{t+1}) = \text{WorldModel}(\mathbf{B}_t, \mathbf{a}^i_t). 
\end{equation}
  
Furthermore, this world model is able to predict the states \(\mathbf{B}_{t+K}^i\) of the future \( K \) steps in a recurrent manner, as shown by the following equation
\begin{equation}\label{eq:recurrent}
  (\mathbf{B}_{t+K}^i, \mathbf{a}^i_{t+K}) = \text{WorldModel}(\mathbf{B}_{t+K-1}^i, \mathbf{a}^i_{t+K-1}).  
\end{equation}
The process above is conducted for every state-action pair in a parallel manner. It is important to note that the size of BEV states is relatively small (e.g., \( h = w = 8 \)), so the world model does not introduce significant computational overhead.
Next, we introduce the Reward Model, which predicts rewards for a trajectory based on its future states.

\subsection{Reward Model}
\label{sec:reward_predictor}
In this section, we first present the reward types and then describe how they are learned.

\paragraph{Reward Types.}
Following~\cite{li2024hydra}, we have two types of rewards: the imitation reward and the simulation reward.
The imitation reward \( r_{\text{im}} \) measures how well the predicted trajectory mimics an expert trajectory.
The detailed implementation will be introduced in Sec.~\ref{sec:bev_space_supervision}.
The simulation rewards \( r_{\text{sim}} \) measures trajectory quality based on simulator-defined criteria.
Following~\cite{navsim}, we assess trajectories using the following five criteria: no collisions (NC), drivable area compliance (DAC), time-to-collision (TTC), comfort (Comf), and ego progress (EP). 
As a result, we have $r_{\text{sim}} = \{r_{\text{sim}}^{\text{NC}}, r_{\text{sim}}^{\text{DAC}}, r_{\text{sim}}^{\text{TTC}}, r_{\text{sim}}^{\text{Comf}},
r_{\text{sim}}^{\text{EP}}\}.
$
Given $r_{\text{im}}$ and $r_{\text{sim}}$, the final reward $r_{\text{final}}$ is defined as 
\begin{equation}\label{eq:score_weighting}
\begin{aligned}
r_{\text{final}} = - \Big( 
    & w_1 \log r_{\text{im}} 
    + w_2 \log r_{\text{sim}}^{\text{NC}} 
    + w_3 \log r_{\text{sim}}^{\text{DAC}} \\
    & + w_4 \log \big( 
        5 r_{\text{sim}}^{\text{TTC}} 
        + 2 r_{\text{sim}}^{\text{Comf}} 
        + 5 r_{\text{sim}}^{\text{EP}} 
    \big) 
\Big),
\end{aligned}
\end{equation}
where $w_i$ are hyper-parameter.

\paragraph{Reward Prediction.}
Our method needs these rewards for trajectory evaluation in the inference time, so we propose reward prediction. Take the $i$-th refined trajectory $\mathbf{\hat{\tau}}_i$ for illustration, the previous approaches~\cite{li2024hydra} typically predict a reward based solely on the current state.
With the help of the world model, our reward model predicts the reward $\mathbf{r_i}$ of $i$-th trajectory based on both the current BEV state and future $K$ steps BEV states. 
This process is illustrated in the following equation
\begin{equation}\label{eq:scorenet}
    \mathbf{r}_i = \text{RewardModel}(\mathbf{B}_{\text{all}}^{i},\mathbf{a}_{\text{all}}^{i}),
\end{equation}
where $ \mathbf{B}_{\text{all}}^{i} = \text{cat}([\mathbf{B}_t,\mathbf{B}_{t+1}^i,...,\mathbf{B}_{t+K}^i])$
and 
$\mathbf{a}_{\text{all}}^{i} = \text{MLP}(\text{cat}([\mathbf{a}_t^i,\mathbf{a}_{t+1}^i,...,\mathbf{a}_{t+K}^i]))
$.
$\text{cat}([\cdot])$ represents the concatenate operation along the channel dimension. 
In detail, Reward Model consists of three parts.
It first applies 2D convolutions to $\mathbf{B}_{\text{all}}^{i}$ to aggregate information from different regions and time steps of the BEV states.
Next, a global average pooling is applied to form $\mathbf{B}_{\text{pool}} \in \mathbb{R}^c$, $c$ is the channel dimension.
We then fed $\mathbf{S}^i = \text{cat}([\mathbf{B}_{\text{pool}}, \mathbf{a}_{\text{all}}^{i}])$, where $\mathbf{S}^i \in \mathbb{R}^{2c}$,
into an MLP head to predict the reward $\mathbf{r}_i \in \mathbb{R}^{m}$, where $m$ denotes the number of reward.
Next, we introduction the definition of these rewards and select a final trajectory based on these rewards.

The refined trajectory with the highest final reward is selected as the final trajectory.
Next, we introduce the supervision of these predicted rewards and BEV states.

\subsection{Supervision in BEV Space} 
\label{sec:bev_space_supervision}
The world model predicts multiple future feature states corresponding to different trajectories. 
However, supervising these predictions in the image space is challenging, as driving logs provide only a single future trajectory. 
To address this, we propose supervising in the BEV space.
Supervising in BEV space alleviates the difficulty of accurately modeling future states in image space while also improving computational efficiency. 
Many real-world traffic simulators support traffic simulations in the BEV space. 
They generate realistic and reliable future BEV scenarios along with rewards. 
We adopt nuPlan~\cite{nuplan} as our traffic simulator for its broad adoption and realistic scenario simulations.

\paragraph{Supervision of BEV States.}
For BEV state supervision, we decode a semantic BEV map based on the BEV state to enable explicit supervision as Figure~\ref{fig:pipeline} (b) shows.
We utilize upsampling and transposed convolution layers to decode a BEV semantic map from the BEV state.
The dimension of the BEV semantic map is \( H \times W \times L \), where \( L \) denotes the number of semantic classes. 
This BEV semantic map provides a comprehensive representation of a BEV scenario, covering classes including background, road, walkways, centerline, static objects, vehicles and pedestrians. 
Each element is represented as a one-hot encoded channel in $L$.
For each time step $t+k$, We use Focal Loss~\cite{lin2017focalloss} to supervise the predicted BEV semantic map $\mathbb{B}_{t+k}$ as 
\begin{equation}
\mathcal{L}_{\text{BEV}} = \text{FocalLoss}(\mathbb{B}_{t+k}, \mathbb{B}^*_{t+k}),    
\end{equation}
where $\mathbb{B}^*_{t+k}$ is the ground truth simulated BEV semantic map at time $t+k$ provided by the simulator.

\paragraph{Supervision of Simulation Rewards.}
Regarding the simulation rewards, we leverage the simulator to produce five rewards for evaluating a trajectory. 
Specifically, given the $i$-th trajectory $\mathbf{\hat{\tau}^i}$, the simulator simulates the future location of the other agents and ego vehicle and produces a simulated future BEV semantic map with agents.
Next, it uses a rule-based evaluator to evaluate this simulated future BEV scenarios for producing the corresponding target simulation rewards of $\mathbf{\hat{\tau}^i}$.
Given the rewards provided by the simulator, we use the Binary Cross Entropy (BCE) loss to supervise the predicted simulation reward $r_{\text{sim}}$.
As a result, the loss is formulated as
\begin{equation}
\mathcal{L}^\text{sim}_{\text{reward}} = \text{BCE}(r_{\text{sim}}, r_{\text{sim}}^*).
\end{equation}

\paragraph{Supervision of Imitation Reward.}\label{sec:supervision_expert}
In addition to the supervision from the simulator, our framework is also guided by the expert driver. 

We compute the target of imitation reward by first calculating the L2 distances $d_i$ between the $i$-th trajectory anchor and an expert trajectory. 
Then, we apply the negative of these distances from $N$ trajectory anchors to the softmax function, yielding
$
r^*_{\text{im}, i} = \frac{\exp(-d_i)}{\sum_{j=1}^N \exp(-d_j)}.
$
Intuitively, a trajectory anchor closer to the expert trajectory than others will receive a higher reward.
As a result, the imitation reward $r^*_{\text{im}} \in \mathbb{R}^N$ quantifies the similarity of each trajectory anchor to the human trajectory. 
The predicted imitation reward $r_{\text{im}}$, which is produced by Eq.~\ref{eq:scorenet}, are supervised using the Cross Entropy loss as 
\begin{equation}
\mathcal{L}^\text{im}_{\text{reward}} = \text{CrossEntropy}(r_{\text{im}}, r^*_{\text{im}}).    
\end{equation}

\paragraph{Supervision of Trajectory.}
For supervising the refined trajectories, we adopt the commonly used winner-take-all strategy~\cite{qcnet}. 
In this strategy, we select the trajectory anchor that is closest to the expert trajectory.
Only the corresponding refined trajectory of this anchor will be supervised.
Let this trajectory anchor be denoted as $\mathbf{\tau}_i$. We use the L1 loss to measure the difference between the expert trajectory $\mathbf{\tau}^*$ and the refined trajectory $\mathbf{\hat{\tau}}_i$ as
\begin{equation}\label{eq:loss_traj}
\mathcal{L}_{\text{traj}} = |\mathbf{\hat{\tau}}_i - \mathbf{\tau}^*|.  
\end{equation}
Instead of supervising all refined trajectories, we focus solely on the most optimal one. This strategy helps the model capture a diverse range of trajectory patterns, with each anchor specializing in a particular trajectory modality.

The overall training loss $\mathcal{L}_{\text{total}}$ is as follow
\begin{equation} 
\mathcal{L}_{\text{total}} = \mathcal{L}_{\text{BEV}} + \mathcal{L}^\text{sim}_{\text{reward}} + \mathcal{L}^\text{im}_{\text{reward}} + \mathcal{L}_{\text{traj}}. 
\end{equation}

\begin{table*}[t!]
    \begin{center}
        \resizebox{0.95\textwidth}{!}{
            \begin{tabular}{l|c|c|c|ccccc|>{\columncolor{gray!30}}c}
                \toprule
                Method & Input & \#Traj. & Traj. Eval. & \textbf{NC} $\uparrow$ & \textbf{DAC} $\uparrow$ & \textbf{EP} $\uparrow$ & \textbf{TTC} $\uparrow$ & \textbf{Comf.} $\uparrow$ & \textbf{PDMS} $\uparrow$ \\ 
                \midrule
                Human & - & - & - & 100 & 100 & 87.5 & 100 & 99.9 & 94.8 \\ 
                \midrule
                Constant Velocity & - & 1 & $\times$ & 69.9 & 58.8 & 49.3 & 49.3 & 100.0 & 21.6 \\ 
                Ego Status MLP & - & 1 & $\times$ & 93.0 & 77.3 & 62.8 & 83.6 & 100.0 & 65.6 \\ 
                VADv2~\citep{paradrive} & C & 8192 & Rule-based & 97.9 & 91.7 & 77.6 & 92.9 & 100.0 & 83.0 \\  
                UniAD~\citep{hu2022uniad} & C & 1 & Rule-based & 97.8 & 91.9 & 78.8 & 92.9 & 100.0 & 83.4 \\ 
                LTF~\citep{transfuser} & C & 1 & $\times$ & 97.4 & 92.8 & 79.0 & 92.4 & 100.0 & 83.8 \\ 
                PARA-Drive~\citep{paradrive} & C & 1 & Rule-based & 97.9 & 92.4 & 79.3 & 93.0 & 99.8 & 84.0 \\ 
                TransFuser~\citep{transfuser} & C \& L & 1 & $\times$ & 97.7 & 92.8 & 79.2 & 92.8 & 100.0 & 84.0 \\
                LAW~\citep{law} & C & 1 & $\times$ & 96.4 & 95.4 & 81.7 & 88.7 & 99.9 & 84.6 \\
                DRAMA~\citep{yuan2024drama} & C \& L  & 1 & $\times$ & 98.0 & 93.1 & 80.1 & 94.8 & 100.0 & 85.5 \\
                Hydra-MDP~\cite{li2024hydra} & C \& L  & 8192 & Model-free & 98.3 & 96.0 & 78.7 & 94.6 & 100.0 & 86.5 \\
                \midrule
                \methodname{} & C \& L  & 256 & Model-based & 98.5 & 96.8 & 81.9 & 94.9 & 99.9 & \cellcolor{gray!30}88.3 \\ 
                \bottomrule
            \end{tabular}
        }
    \caption{\textbf{Comparison with the SOTA approaches on NAVSIM test set.} 
    Rule-based: UniAD and PARA-Drive use occupancy maps, while VADv2 relies on vectorized maps for trajectory evaluation following specific rules.
    Model-free: Hydra-MDP evaluates trajectory without world modeling.
    Model-based: our \methodname{} evaluates trajectory with the assistance of the BEV world model.
    NC: no at-fault collision.
    DAC: drivable area compliance.
    EP: ego progress.
    TTC: time-to-collision.
    Comf.: comfort.
    PDMS: the predictive driver model score.
    Traj.: Trajectory.
    Eval.: Evaluation.
    C: Camera.
    L: LiDAR.
    \label{tab:sota_navsim}
    }
    \end{center}
\end{table*}

\begin{table}[]
\centering
\resizebox{\columnwidth}{!}{%
\begin{tabular}{l|c|c>{\columncolor{gray!30}}c}
\toprule
Method & Traj. Eval.     & Success Rate$\uparrow$ & Driving Score$\uparrow$ \\ \midrule
AD-MLP~\cite{zhai2023rethinking} & -               & 0.00         & 18.05         \\
UniAD~\cite{hu2022uniad}  & Rule-based      & 16.36        & 45.81         \\
VAD~\cite{jiang2023vad}    & Rule-based & 15.00        & 42.35         \\
TCP~\cite{wu2022tcp}    & -               & 30.00        & 59.90         \\
\methodname{}  & Model-based      & 31.36        & 61.71         \\ \bottomrule
\end{tabular}%
}
\caption{\textbf{Comparison with SOTA approaches on closed-loop Bench2Drive~\cite{jia2024bench2drive} benchmark on CARLA simulator.}
Traj.: Trajectory.
Eval.: Evaluation.
}
\vspace{-3mm}
\label{tab:carla_sota}
\end{table}

\section{Experiments}
\subsection{Benchmark}
\noindent\textbf{NAVSIM Dataset} The NAVSIM dataset \citep{navsim} is constructed based on nuPlan \citep{nuplan}. Specifically, OpenScene \citep{openscene} first down-sampled the nuPlan data from 10Hz to 2Hz to condense it into 120 hours of driving logs. NAVSIM resampled the data from OpenScene to emphasize challenging scenarios, reducing simple situations like straight-line driving. 
The dataset is divided into two parts: Navtrain and Navtest, comprising 1192 scenarios for training and validation and 136 scenarios for testing.
Unlike nuScenes~\cite{nuscenes19}, which collects data in simpler, slower driving scenarios, NAVSIM ensures that challenging scenarios are prioritized, making simply fitting ego status insufficient for planning.

\noindent\textbf{NAVSIM Metrics} The original end-to-end driving metrics were primarily designed to evaluate the deviation between the predicted trajectories and those driven by human experts. However, this evaluation approach has proven inadequate, resulting in poor performance in real-world driving contexts. NAVSIM introduces more practical and reliable metrics, focusing on aligning open-loop and closed-loop metrics.
Specifically, NAVSIM evaluates model performance using the Predictive Driver Model Score (PDMS), which is calculated based on five factors: No At-Fault Collision (NC), Drivable Area Compliance (DAC), Time-to-Collision (TTC), Comfort (Comf.), and Ego Progress (EP). 
The PDMS is calculated as
$
    \text{PDMS} = \text{NC} \times \text{DAC} \times \frac{5 \times \text{EP} + 5 \times \text{TTC} + 2 \times \text{C}}{12}.
$

\noindent\textbf{Bench2Drive Dataset and Metrics}
To further evaluate our framework, we conduct closed-loop experiments in the CARLA simulator~\cite{dosovitskiy2017carla}. Specifically, we adopt the recently proposed Bench2Drive~\cite{jia2024bench2drive} benchmark as our closed-loop evaluation benchmark. Bench2Drive provides a standardized training dataset on CARLA, ensuring fair comparisons across different methods.
The benchmark consists of 220 short evaluation routes (around 150 meters each) distributed across all CARLA towns, with each route containing one safety-critical scenario. 
This design effectively reduces the evaluation variance.  
For metric, it utilizes the standard CARLA metric, Driving Score (DS), as the primary metric.
Additionally, Bench2Drive reports the success rate, which represents the proportion of successfully completed routes. 
A route is considered successful if and only if DS reaches 100\% on this route.

\subsection{Implementation Details}
\noindent\textbf{NAVSIM}
For input data, it is aligned with TransFuser~\cite{transfuser}.
We concatenate the front-view image with center-cropped front-left and front-right images, resulting in a combined resolution of $256 \times 1024$ pixels. For LiDAR, the $64m \times 64m$ point cloud surrounding the ego vehicles is used.
For network architecture, we employ ResNet34 as the backbone for BEV feature extraction following TransFuser~\cite{transfuser}.
The world model consists of two transformer decoder layers.
We use $N=256$ trajectory anchors.
The reward weights are set as follows: $w_1 = 0.1, w_2 = w_3 = 0.5, w_4=1.0$.
The training is conducted on the Navtrain split using 8 NVIDIA L20 GPUs with a total batch size of 128, distributed across 30 epochs. 
In the ablation study section, we train our model for 20 epochs instead of 30 to accelerate the training process.
For fast training, we pre-compute the simulation results (BEV semantic maps and scores) based on the trajectory anchors. 
We then feed trajectory anchors into the trajectory evaluation module during training.
During testing, the trajectory evaluation module evaluates the refined trajectory instead of the anchor trajectories.
We utilize the Adam optimizer with a learning rate of 1e-4.

\noindent\textbf{Bench2Drive}
For Bench2Drive, we adopt TCP~\cite{wu2022tcp} as our baseline and integrate the trajectory evaluation module. We choose TCP as it is the best open-source framework officially provided by Bench2Drive. Similar to NAVSIM, we utilize imitation rewards along with simulation-based rewards, including No Collision (NC) and Drivable Area Compliance (DAC). The number of trajectory anchors is set to 256.
Notably, rather than employing the winner-take-all strategy (Eq.~\ref{eq:loss_traj}) to supervise only the best trajectory, we find that supervising all trajectories leads to improved performance. The model is trained for 27 epochs with a batch size of 300. We use the Adam optimizer with a learning rate of 1e-4.

\subsection{Comparison with SOTA}
\noindent\textbf{NASIM}
Table~\ref{tab:sota_navsim} compares our method with state-of-the-art approaches on the NAVSIM test set. By leveraging the world model to simulate future scenarios and guide trajectory evaluation, our method achieves a PDMS of 87.1. \methodname{} outperforms the previous model-free approach, Hydra-MDP, highlighting the advantages of model-based trajectory evaluation.

\noindent\textbf{Bench2Drive}
To further validate our approach, we conduct a closed-loop evaluation on the Bench2Drive benchmark within the CARLA simulator. As shown in Table~\ref{tab:carla_sota}, our method improves the Driving Score by 1.81 points, demonstrating its effectiveness in various evaluation settings.

\subsection{Ablation Studies}
\noindent\textbf{Trajectory evaluation and future state prediction matter.}
In this experiment, we investigate the importance of the trajectory evaluation and future state prediction as  Table~\ref{tab:score_matters} shows.
The first row in Table \ref{tab:score_matters} represents our baseline TransFuser~\cite{transfuser}, which performs trajectory prediction only.
In the second row, we change to our framework, which consists of both trajectory prediction and trajectory evaluation.
However, we do not used the predicted future states as the input of Reward Model.
That is, $\mathbf{B}_{\text{all}} = \mathbf{B}_t$ and $\mathbf{a}_{\text{all}} = \mathbf{a}_t$ in Eq.~\ref{eq:scorenet}.
In the third row, we integrate the BEV world model to predict future states, leading to notable performance improvements across multiple metrics.
\begin{table}[!]
    \begin{center}
        \resizebox{\columnwidth}{!}{
            \begin{tabular}{c|c|ccccc|>{\columncolor{gray!30}}c}
                \toprule
                Traj. Eval. & Future States & \textbf{NC} $\uparrow$ & \textbf{DAC} $\uparrow$ & \textbf{EP} $\uparrow$ & \textbf{TTC} $\uparrow$ & \textbf{Comf.} $\uparrow$ & \textbf{PDMS} $\uparrow$ \\ 
                \midrule
                $\times$ & $\times$ & 96.4 & 91.5 & 76.2 & 90.3 & 99.9 & 81.0 \\ 
                $\checkmark$ & $\times$ & 97.1 & 92.9 & 78.2 & 91.9 & 100.0 & 83.2 \\ 
                $\checkmark$ & $\checkmark$ & 98.0 & 94.7 & 79.9 & 93.4 & 100.0 & 85.6  \\ 
                \bottomrule
            \end{tabular}
    }
    \caption{\textbf{Ablation study on trajectory evaluation and future state prediction.} 
    Traj: Trajectory.
    Eval.: Evaluation.
    }
    \label{tab:score_matters}
    \end{center}
\end{table}

\noindent\textbf{Imitation and simulation rewards are complementary.}  
To investigate how each type of reward affects driving performance, we conduct an ablation study, as shown in Table~\ref{tab:abl_rewards}. 
During inference, we adjust the reward weights in Eq.~\ref{eq:score_weighting} to isolate their effects: setting \( w_1 = 0 \) enables evaluation with only simulation rewards, while setting \( w_2 = w_3 = w_4 = 0 \) tests performance with only an imitation reward.  
The results indicate that imitation and simulation rewards emphasize different aspects of planning. 
Trajectory evaluation using imitation rewards performs better in NC and TTC, while using simulation rewards excels in DAC and EP.
By integrating both, the model leverages their respective strengths, achieving improved overall performance across all key metrics.
\begin{table}[!]
    \begin{center}
        \resizebox{\columnwidth}{!}{
            \begin{tabular}{c|c|ccccc|>{\columncolor{gray!30}}c}
                \toprule
                Imi. Reward & Sim. Rewards & \textbf{NC} $\uparrow$ & \textbf{DAC} $\uparrow$ & \textbf{EP} $\uparrow$ & \textbf{TTC} $\uparrow$ & \textbf{Comf.} $\uparrow$ & \textbf{PDMS} $\uparrow$ \\ 
                \midrule
                $\checkmark$ & $\times$ & 97.9 & 92.5 & 77.7 & 93.4 & 99.9 & 83.5 \\ 
                $\times$ & $\checkmark$ & 96.5 & 94.7 & 79.2 & 90.1 & 99.4 & 83.6 \\ 
                $\checkmark$ & $\checkmark$ & 98.0 & 94.7 & 79.9 & 93.4 & 100.0 & 85.6  \\ 
                \bottomrule
            \end{tabular}
    }
    \caption{\textbf{Ablation study on the imitation and simulation rewards.} The experiment shows that imitation and simulation rewards are complementary.
    Imi: Imitation.
    Sim.: Simulation.
    }
    \label{tab:abl_rewards}
    \end{center}
\end{table}

\noindent\textbf{Recurrent future state prediction helps trajectory evaluation.}
Our framework supports predicting future states recurrently, as detailed in Eq.~\ref{eq:recurrent}.
In Table~\ref{tab:used_time_steps}, we present the results obtained by varying the number of BEV states predicted by the world model.
Our findings indicate that using a finer time step for future state prediction significantly improves performance.
This is because a finer time step provides richer temporal information, which aids trajectory evaluation.
\begin{table}[!]
    \begin{center}
        \resizebox{\columnwidth}{!}{
            \begin{tabular}{l|ccccc|>{\columncolor{gray!30}}c}
                \toprule
                 Future Prediction Steps & \textbf{NC} $\uparrow$ & \textbf{DAC} $\uparrow$ & \textbf{EP} $\uparrow$ & \textbf{TTC} $\uparrow$ & \textbf{Comf.} $\uparrow$ & \textbf{PDMS} $\uparrow$ \\ 
                \midrule
                0s $\rightarrow$ 4s & 97.4 & 93.6 & 79.1 & 91.9 & 100.0 & 84.0 \\ 
                 0s $\rightarrow$ 2s $\rightarrow$ 4s & 98.0 & 94.7 & 79.9 & 93.4 & 100.0 & 85.6 \\ 
                \bottomrule
            \end{tabular}
        }
    \caption{\textbf{Ablation study on the different number of time steps in the future prediction.} Our BEV world model predicts future states in a recurrent manner.
    A $\rightarrow$ B: the predicted future state at time B is based on the state at time A.
    }
    \label{tab:used_time_steps}
    \end{center}
\end{table}

\noindent\textbf{More trajectories help .}
\begin{table}[!]
    \begin{center}
        \resizebox{\columnwidth}{!}{
            \begin{tabular}{c|ccccc|>{\columncolor{gray!30}}c|c}
                \toprule
                \#Trajs. & \textbf{NC} $\uparrow$ & \textbf{DAC} $\uparrow$ & \textbf{EP} $\uparrow$ & \textbf{TTC} $\uparrow$ & \textbf{Comf.} $\uparrow$ & \textbf{PDMS} $\uparrow$ & \textbf{Latency} $\downarrow$\\
                \midrule
                64  & 97.4 & 94.1 & 79.6 & 91.9 & 100.0 & 84.5 & 17.2 ms \\
                128 & 97.9 & 94.6 & 80.0 & 93.0 & 100.0 & 85.4 & 17.9 ms \\
                256 & 98.0 & 94.7 & 79.9 & 93.4 & 100.0 & 85.6 & 18.7 ms \\
                \bottomrule
            \end{tabular}
        }
    \caption{\textbf{Ablation study on the different number of trajectories.} 
    The number of trajectories is the same during training and testing.    
    The latency is evaluated on an NVIDIA L20 GPU.
    Trajs.: Trajectories.}
    \label{tab:num_trajs}
    \end{center}
\end{table}

This ablation study explores the effect of varying the number of trajectories on overall performance. 
Specifically, the model in each row is trained and tested under the same number of trajectories, and the world model is fed with refined trajectories.
As shown in Table \ref{tab:num_trajs}, using a small number of trajectories, such as 64, significantly degrades performance. 
As the number of trajectories increases, performance improves notably, with a marked improvement observed when increasing from 64 to 128 trajectories. 
However, as the number of trajectories increases from 128 to 256, the performance gains diminish, indicating that the performance of the model approaches its optimal level.
Based on these results, we choose 256 as the default number of trajectories for our experiments.

\noindent\textbf{Latency analysis.}
A common concern is the overall latency of our framework.
To address this, we measured its latency on an NVIDIA L20 GPU, as shown in Table~\ref{tab:num_trajs}.
Since the GPU processes state-action pairs in parallel, our framework remains efficient even as the number of trajectories increases.
The total latency is only 18.7 ms, well within the real-time requirements for end-to-end autonomous driving.

\vspace{3mm}
\noindent\textbf{Generalization ability across different trajectories.}
As shown in Table~\ref{tab:unseen_trajs}, we investigate the generalization ability of our trajectory evaluation module.
To be specific, we sample $N=1024$ instead of $N=256$ trajectory anchors from the dataset.
Since our model is trained under the setting of $N=256$ trajectory anchors, the model has never seen these $N=1024$ trajectories.
Surprisingly, we find a $1.3$ PDMS increase when using these unseen trajectory anchors.
Besides, the model performs even better when using refined trajectories as inputs. 
These refined trajectories continuously change based on varying sensor inputs.
This shows that the proposed world model has great generalization ability when dealing with unseen trajectories.
\begin{table}[!]
    \begin{center}
        \resizebox{\columnwidth}{!}{
            \begin{tabular}{c|c|c|ccccc|>{\columncolor{gray!30}}c}
                \toprule
                \#Train Traj. & \#Test Traj. & Traj. Refine\dag & \textbf{NC} $\uparrow$ & \textbf{DAC} $\uparrow$ & \textbf{EP} $\uparrow$ & \textbf{TTC} $\uparrow$ & \textbf{Comf.} $\uparrow$ & \textbf{PDMS} $\uparrow$ \\
                \midrule
                256 & 256  & $\times$ & 96.6 & 94.6 & 79.0 & 91.1 & 100.0 & 84.0 \\ 
                256 & 1024 & $\times$ & 97.2 & 95.6 & 79.3 & 92.4 & 100.0 & 85.3 \\ 
                256 & 256  & $\checkmark$ & 98.0 & 94.7 & 79.9 & 93.4 & 100.0 & 85.6 \\
                \bottomrule
            \end{tabular}
        }
    \caption{\textbf{Generalization of the trajectory evaluation module across different trajectory configurations.
    } 
    \dag: indicating whether the input trajectories of BEV world model are refined.
    Traj.: Trajectory.}
    \label{tab:unseen_trajs}
    \end{center}
\end{table}

\subsection{Visual Analysis}
\begin{figure}[!ht]
    \centering
    \includegraphics[width=\linewidth]{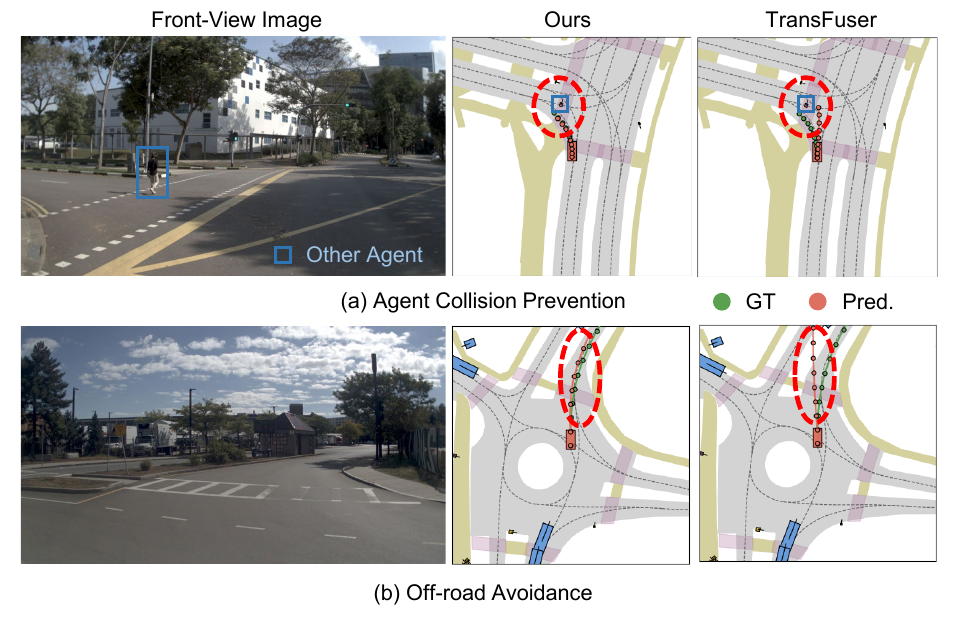}
    \caption{\textbf{Visualization of end-to-end planning trajectories.} We compare our method with TransFuser~\cite{transfuser}, which serves as our trajectory prediction baseline without a trajectory evaluation module. Our trajectory evaluation module effectively filters out low-quality trajectories and helps avoid collisions.}
    \label{fig:vis_main}
\end{figure}

\noindent\textbf{End-to-end planning trajectories.} 
To highlight the improvements in end-to-end driving achieved through trajectory evaluation, we present visualizations of end-to-end planning trajectories in Figure~\ref{fig:vis_main}. 

\begin{figure}[h]
    \centering
    \includegraphics[width=\linewidth]{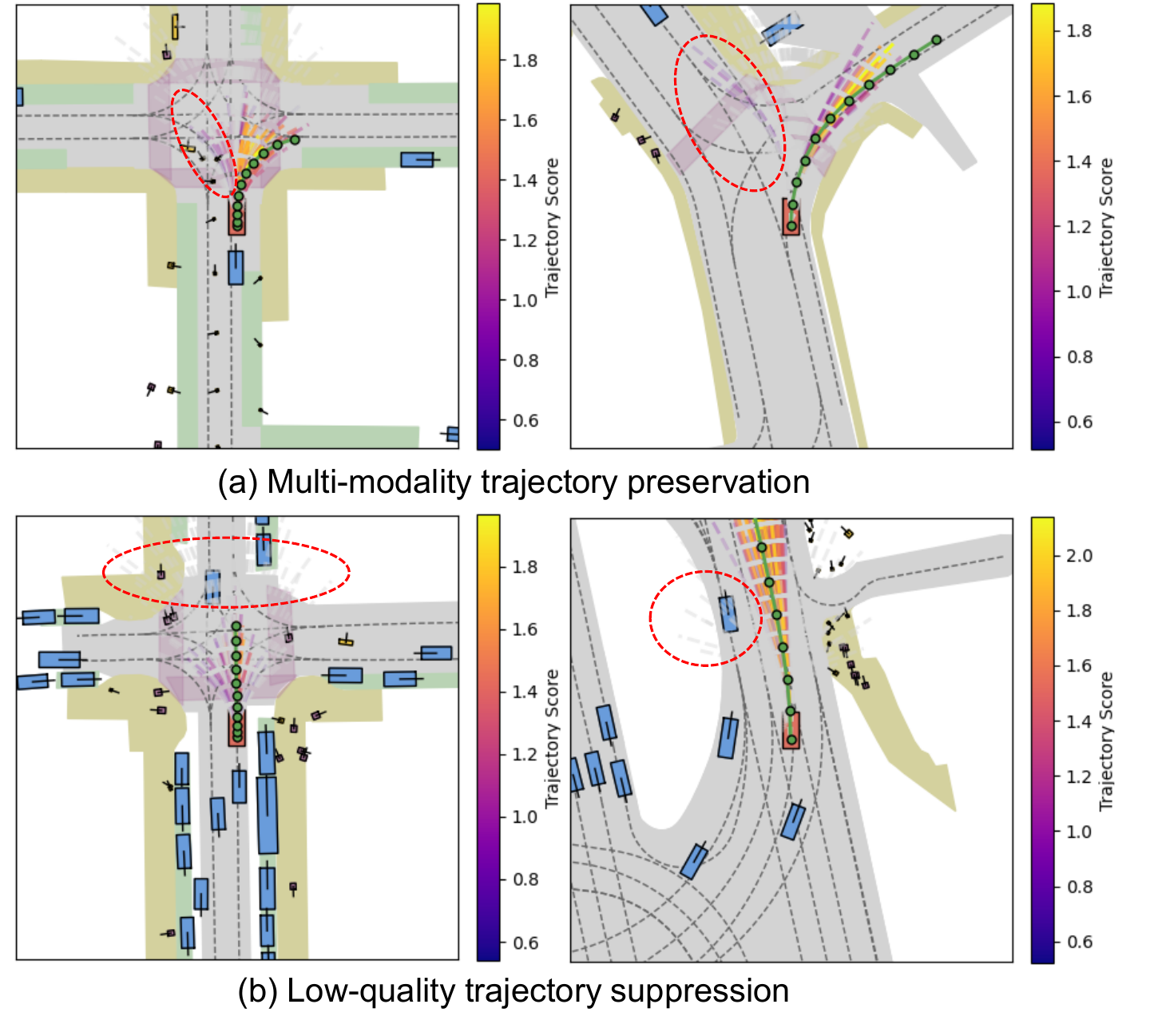}
    \caption{\textbf{Visualization of rewards of all trajectories.}
    Brighter colors indicate higher trajectory rewards, while \textcolor{gray}{gray} trajectories correspond to those with very low rewards.
    (a): \methodname{} effectively retains diverse multi-modal trajectories, as long as they comply with traffic rules and do not result in unsafe outcomes.
    (b): it demonstrates that trajectories considered unreasonable receive significantly lower rewards, highlighting the effectiveness of our trajectory evaluation module.}
    \vspace{-2mm}
    \label{fig:vis_score}
\end{figure}

\vspace{3mm}
\noindent\textbf{Trajectory rewards.}
Figure~\ref{fig:vis_score} illustrates the predicted rewards for different trajectories.

\section{Conclusion} 
In this paper, we introduced a novel framework that leverages a BEV world model for end-to-end trajectory evaluation in autonomous driving. 
By integrating a BEV world model, we enable a more informed evaluation process that considers the dynamic evolution of driving scenarios, leading to more effective trajectory evaluation.
our approach benefits from dense supervision provided by BEV-space traffic simulators, which supply both semantic future states and rule-based reward targets. 
Our method achieves state-of-the-art performance on the NAVSIM and Bench2Drive benchmarks while maintaining real-time efficiency. 
Overall, our work establishes trajectory evaluation as an important research direction for end-to-end autonomous driving. 
We hope our framework serves as a strong baseline and inspires further research in end-to-end online trajectory evaluation.

{
    \small
    \bibliographystyle{ieeenat_fullname}
    \bibliography{main}
}


\end{document}